\documentclass[letterpaper, 10 pt, conference]{ieeeconf}  

\IEEEoverridecommandlockouts                              

\usepackage{amsmath} 
\usepackage{float}
\usepackage{amssymb}
\usepackage{subfigure}
\usepackage{enumerate}
\usepackage[export]{adjustbox}
\usepackage{hyperref}
\usepackage{footnote}
\graphicspath{{images/}}
\usepackage{color}
\usepackage{graphicx}

\title{\LARGE \bf
Robust Navigation for Racing Drones based on \linebreak Imitation Learning and Modularization
}

\author{Tianqi Wang$^{1}$ and Dong Eui Chang$^{1}$
\thanks{$^{1}$Tianqi Wang and Dong Eui Chang are with the Control Laboratory, School of Electrical Engineering, Korea Advanced Institute of Science and Technology, Republic of Korea, {\tt\small tianqi\_wang@kaist.ac.kr; dechang@kaist.ac.kr}}%
\thanks{Experiment video is available at: \url{https://youtu.be/-n_liOxU7rA}}%
\thanks{Open-source code is available at: \url{https://github.com/tianqi-wang1996/DeepRobustDroneRacing}}%
}

\begin{document}

\pdfsuppresswarningpagegroup=1

\maketitle
\thispagestyle{empty}
\pagestyle{empty}

\begin{abstract}
  This paper presents a vision-based modularized drone racing navigation system that uses a customized convolutional neural network (CNN) for the perception module to produce high-level navigation commands and then leverages a state-of-the-art planner and controller to generate low-level control commands, thus exploiting the advantages of both data-based and model-based approaches. Unlike the state-of-the-art method, which only takes the current camera image as the CNN input, we further add the latest three estimated drone states as part of the inputs. Our method outperforms the state-of-the-art method in various track layouts and offers two switchable navigation behaviors with a single trained network. The CNN-based perception module is trained to imitate an expert policy that automatically generates ground truth navigation commands based on the pre-computed global trajectories. Owing to the extensive randomization and our modified dataset aggregation (DAgger) policy during data collection, our navigation system, which is purely trained in simulation with synthetic textures, successfully operates in environments with randomly-chosen photo-realistic textures without further fine-tuning. 
\end{abstract}

\section{INTRODUCTION}
Racing drones aim to traverse tracks composed of arbitrarily placed gates as fast as possible without collision. Designing a fully autonomous racing drone is challenging and requires research on various topics such as state estimation, localization and mapping, path planning, modeling of dynamics, and control. As a result, autonomous drone racing has recently aroused significant research interest to use drones as experiment platforms to develop techniques that could transfer to other robotics applications. 

A possible method for autonomous drone racing is to follow a precomputed global trajectory. However, this method has apparent limitations such as the need for knowing the track layout in advance and the demand of highly accurate state estimation of drones, which still cannot be sufficiently satisfied by the current methods (\cite{vio1}$-$\cite{slam2}).

\begin{figure}[t]
	\centerline{\includegraphics[width=8.5cm]{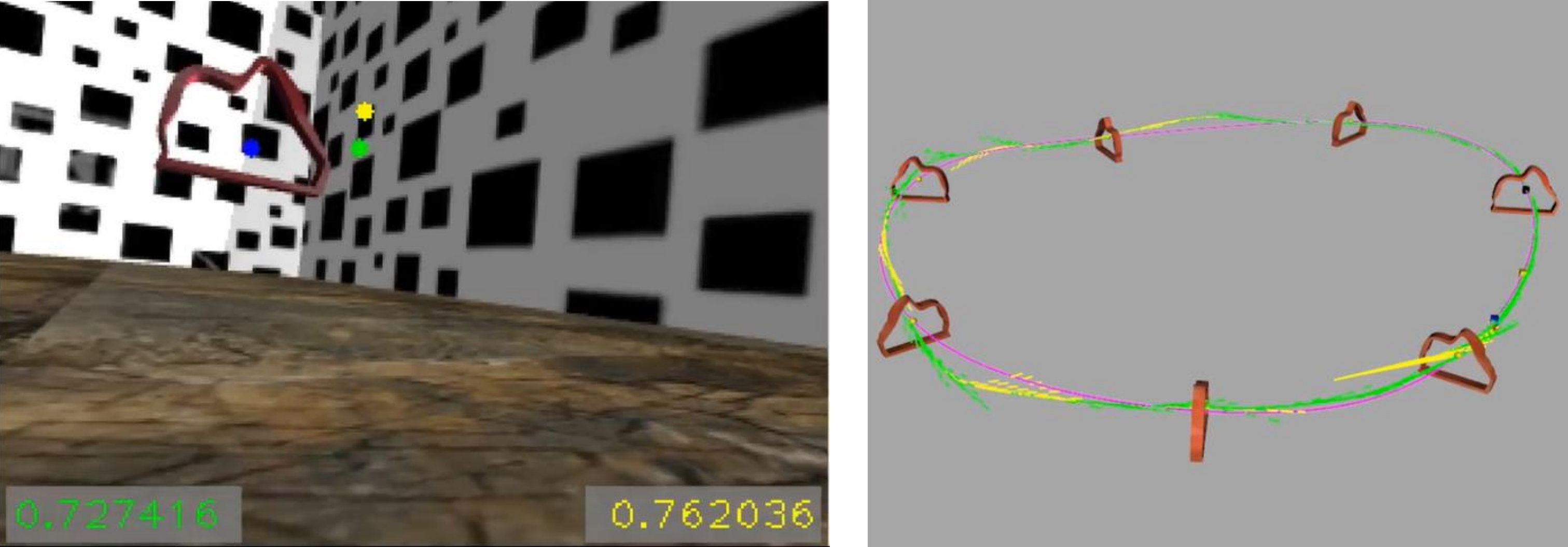}}
	\caption[Our modified DAgger (Dataset Aggregation) policy during data collection.]{Left image - an example of our collected data, the yellow point and number: ground truth navigation direction and normalized navigation speed generated by the expert policy, the green point and number: partially-trained neural network's outputs for the navigation direction and normalized navigation speed, blue point: ground truth of the next gate's center point (2D normalized image coordinate). Right image - illustration of our modified DAgger (Dataset Aggregation) policy during data collection, which normally uses the partially trained network (green trajectories in the right image) to maneuver the drone but switches to the expert policy (yellow trajectories) to recover the drone when it flies far away from the global trajectory (purple trajectory).} \label{fig-intro_DAgger}
\end{figure}
In recent years, many data-driven approaches have been proposed for autonomous navigation (\cite{cite-21}$-$\cite{DNN}). These methods have shown several interesting advantages, such as robustness against drift in state estimation \cite{cite-21, cite-22}, and the ability to learn from failures \cite{cite-24}. Most of the approaches leverage supervised learning \cite{cite-21, cite-22} to imitate expert policy such as human drivers' behavior, in which a significant amount of annotated training data is required. To overcome this problem, Loquercio \textit{et al.} \cite{cite-26} proposed to use datasets for ground vehicles to train autonomous drones instead and Gandhi \textit{et al.} \cite{cite-27} developed a method that automatically generates annotated training data. Besides that, reinforcement learning has also become a popular choice for autonomous navigation \cite{rl1, rl2, rl3}. However, a common limitation of these reinforcement learning-based approaches is that it may take a prohibitively long time to converge when the possible state and action space is too large. Moreover, the trained policy's behavior is very difficult to verify or explain, which makes it risky and unreliable to implement in real platforms like cars and drones.

The methods proposed in \cite{sim2real_prev, sim2real} provide the most insights for this paper. They separate the whole navigation system into perception, planning, and control parts, thus combining the robust perceptual awareness provided by the convolutional neural networks (CNN) with the precision offered by the state-of-art planners and controllers. In comparison to the method in \cite{sim2real_prev}, the authors in \cite{sim2real} added domain randomization during collecting training data, which randomizes the visual scenes such as the floor and background textures. This randomization has led to improved closed-loop performance when evaluated in unseen environments.

In this paper, inspired by \cite{sim2real_prev, sim2real}, we augment the estimated drone states along with the forward-facing camera image as the neural network inputs and design a customized neural network architecture along with our data collection strategy (see Fig. \ref{fig-intro_DAgger}) for training the perception module. The idea proposed in \cite{sim2real_prev, sim2real} to identify waypoints in local body-frame is kept in this paper since it eliminates the problem of state estimation drift and simultaneously enables the drone to navigate through dynamic environments, which can be seen as the main contribution of \cite{sim2real_prev, sim2real}. Besides the visual scene randomization implemented in \cite{sim2real}, we also randomize the track layout and the aggressiveness (maximum velocity) of the drone behavior for every new experiment, making our trained network more robust to condition changes. Our whole system has shown better overall performance in various track layouts in comparison to \cite{sim2real}. Besides, our trained network can generate predictions of the next gate's position and orientation, which makes our perception module's decisions more explainable. 

\section{METHOD}
\subsection{Modularization of the Navigation System}
For the autonomous navigation task, learning an end-to-end policy mapping directly from camera image and drone state inputs to low-level control commands has shown to be unstable and not able to generalize well to unseen environments. Instead, we follow the modularization framework proposed in \cite{sim2real} to separate the system into perception, planning, and control module. In this paper, we focus on developing a robust CNN-based perception module that produces desired navigation direction and navigation speed from the camera image and drone states, from which  the 3D goal point can be further derived. After that, the planning module calculates a minimum-jerk trajectory \cite{minijerk} to the goal point, which is then tracked by the control module \cite{mpc_tracker}. 

\begin{figure*}[t]
	\centerline{\includegraphics[width=14cm]{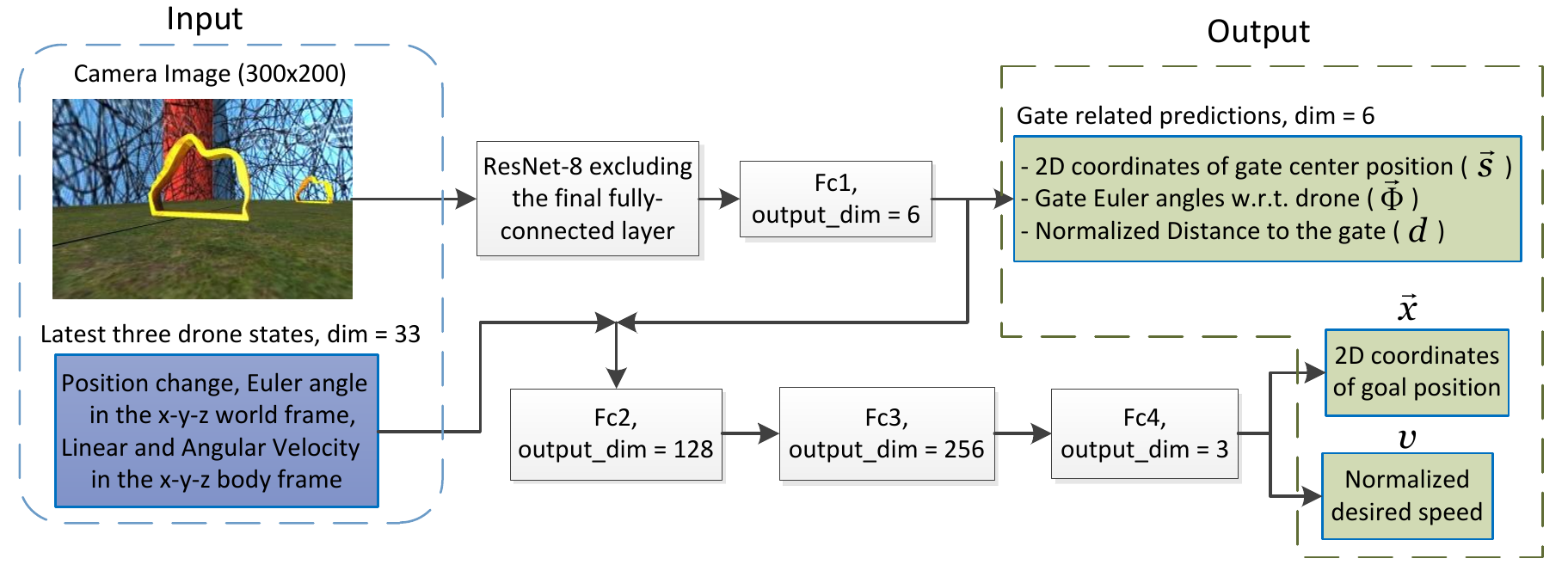}}
	\caption[Customized neural network architecture.]{Architecture of our customized neural network (Fc stands for fully connected layer). Use ReLU as the activation layer, but omitted in the above figure for brevity.} 
	\label{fig-neural_network}
\end{figure*}

\textit{Perception module}: 
Instead of only relying on the current camera image as proposed in \cite{sim2real}, we augment it with the latest three estimated drone states as the CNN inputs. We have also designed a customized convolutional neural network which takes the aforementioned inputs to generate the navigation direction $\vec{x} \in \left[-1,1\right]^2$ (in 2D normalized image coordinates) and normalized navigation speed $v \in \left[0,1\right]$. The details of our customized network architecture are shown in Fig. \ref{fig-neural_network}. 

\textit{Planning module}: Back-projecting the 2D normalized image coordinates $\vec{x}$ generates a camera projection ray on which the 3D goal point is located. We find the 3D goal point on the projection ray at a distance that is proportional to the normalized navigation speed $v$. We call this distance the planning length ($\ell_\textrm{train}$ for training and $\ell_\textrm{test}$ for testing, see Eq. \ref{eq-l_train}, Eq. \ref{eq-l_test}). Besides, the desired navigation speed $v_\textrm{des}$ can be calculated by $v_\textrm{des} = v_\textrm{max}\,\cdot\,v$, where $v_\textrm{max}$ is the maximum flying speed defined by the user such that the aggressiveness of the flying behavior can be changed with a single trained neural network. After obtaining the 3D goal point and the desired navigation speed $v_\textrm{des}$,  we then use the minimum-jerk trajectory \cite{minijerk} to generate the local trajectory $\mathcal{T}_\ell$.

\textit{Control module}: The local trajectory $\mathcal{T}_\ell$ is tracked by a controller proposed in \cite{mpc_tracker}, which takes the rotor drag effect of the drone dynamics into account. Note that, the local trajectory $\mathcal{T}_\ell$ is planned and tracked in a receding horizon fashion in the sense that it gets re-planned whenever a new prediction from the perception module is available. 

\subsection{Expert Policy}
\label{section-expert_policy}
\begin{figure}[t]
	\setlength{\belowcaptionskip}{-0.3cm} 
	\centerline{\includegraphics[width=8cm]{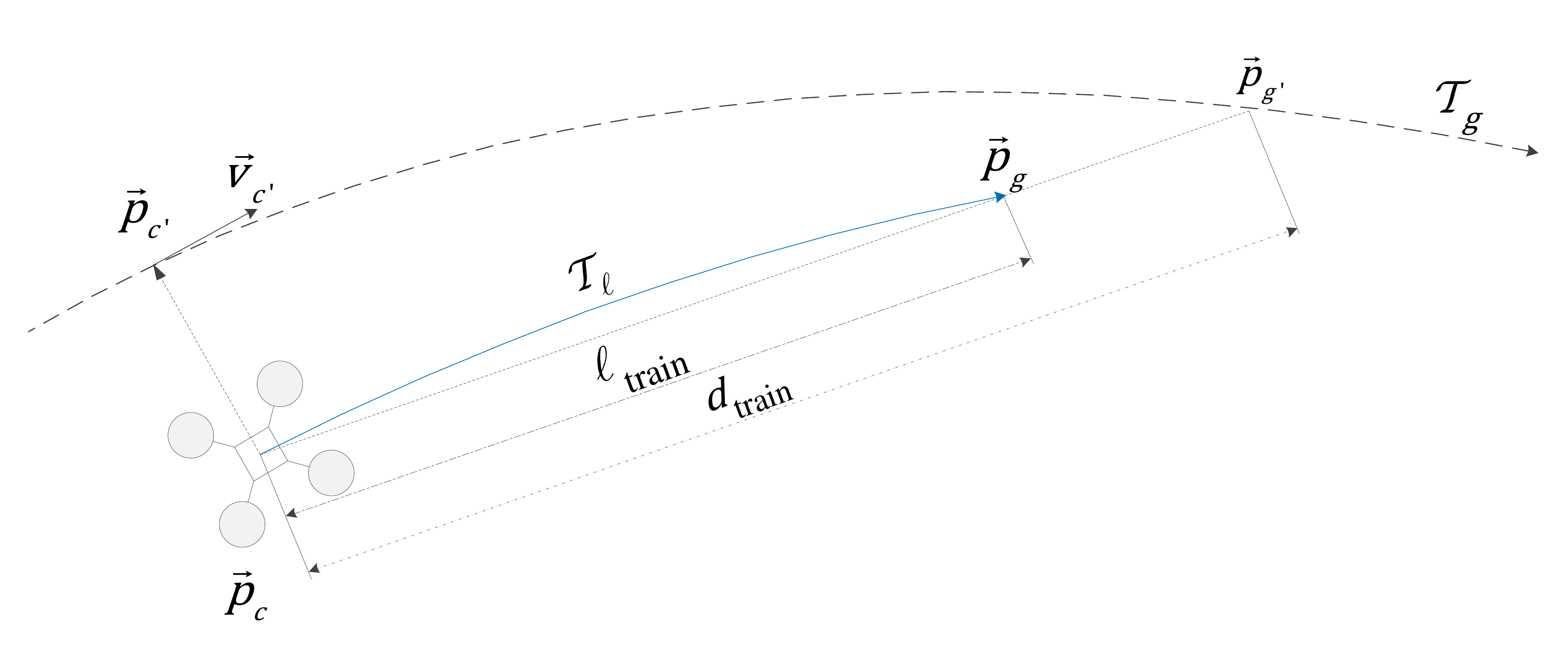}}
	\caption[Training data generation procedure.]{Illustration of the expert policy. $\mathcal{T}_g$: global trajectory, $\mathcal{T}_l$: local trajectory,  $\vec{p}_g$: 3D goal point. $\vec{p}_c$: drone position, $\vec{p}_{c^{'}}$ and $\vec{v}_{c^{'}}$: the closest point on $\mathcal{T}_g$ and its velocity, $\vec{p}_{g^{'}}$: the point that generates ground truth navigation direction. $d_\textrm{train}$ and $l_\textrm{train}$: prediction horizon and planning length at training time.} \label{fig-expert_policy}
\end{figure}

Given the pre-computed global trajectory and accurate positions of the drone and the gates, the ground truth labels for the navigation direction and normalized navigation speed can be automatically generated by our expert policy, which is similar to the one proposed in \cite{sim2real} with the difference in the selection of the 3D goal point. We then use imitation learning to train our customized neural network, which serves as the perception module. Note that, at testing time, only estimated drone states and the current camera image are required to navigate. 

\textit{Expert policy}: The procedure of our expert policy is demonstrated in Fig. \ref{fig-expert_policy}. Using the minimum-snap trajectory implementation from \cite{minisnap}, we first generate a global trajectory $\mathcal{T}_g$ that passes through all gates of the track. Given the current drone position $\vec{p}_c \in \mathbb{R}^{3}$, we define the prediction horizon $d_\textrm{train} \in \mathbb{R}$ as follow:
\begin{equation}
d_\textrm{train} = \textrm{min}(d_\textrm{max}, \textrm{max}(d_\textrm{min}, \textrm{min}(d_\textrm{prev}, d_\textrm{next}))),
\label{eq-dtrain}
\end{equation}
\noindent where $d_\textrm{prev}$ and $d_\textrm{next}$ are the distances from the drone to the previous gate and the next gate, $d_\textrm{min}$ and $d_\textrm{max}$ are the minimum and maximum prediction horizon specified by users, which we set as 1.0 m and 10.0 m respectively. 

A point $\vec{p}_{g^{'}} \in \mathbb{R}^{3}$ on the global trajectory $\mathcal{T}_g$, which is $d_\textrm{train}$ away from the current drone position $\vec{p}_c$ in the forward direction of $\mathcal{T}_g$, can be found. Projecting $\vec{p}_{g^{'}}$ onto the image plane of the camera produces the ground truth navigation direction $\vec{x}_g \in \left[-1,1\right]^2$ (normalized image coordinates). In addition, the closest point $\vec{p}_{c^{'}} \in \mathbb{R}^{3}$ to the drone on $\mathcal{T}_g$ can be found and the ground truth normalized navigation speed $v_g \in \left[0,1\right]$ is calculated by dividing the speed of $\mathcal{T}_g$ at $\vec{p}_{c^{'}}$ with the maximum speed achieved along $\mathcal{T}_g$. After obtaining $\vec{x}_g$ and $v_g$, we then back-project the 2D normalized image coordinates $\vec{x}_g$ to get the camera projection ray and find the 3D goal position $\vec{p}_g$ on it at a distance equal to the planning length at training time $\ell_\textrm{train}$, which is defined proportional to $v_g$:
\begin{equation}
\ell_\textrm{train} = \textrm{min}(\ell_\textrm{max}, \textrm{max}(\ell_\textrm{min}, m_\ell \cdot v_g))
\label{eq-l_train}
\end{equation}
\noindent where $\ell_\textrm{min}$ and $\ell_\textrm{max}$ are user-specified minimum and maximum planning length, and  $m_\ell$ is also a user-specified coefficient. In our case, we choose $\ell_\textrm{min} = $ 0 m,  $\ell_\textrm{max} = $ 3.0 m, $m_\ell = $ 5.0.

\subsection{Customized Neural Network for Navigation}
The expert policy mentioned in \ref{section-expert_policy} is based on the pre-computed global trajectory and accurate poses of the drone and all the gates, which are not available during testing time. However, other sensor data collected while the drone being maneuvered by the expert policy, which are available during testing, such as the camera images and the drone state, can also give us the rationales for the expert policy's decisions. 

In \cite{sim2real}, the authors design their network mapping from the current camera image to the navigation direction and speed generated by the expert policy. However, purely relying on the camera image might hinder the overall navigation performance since the drone can be in many possible states even with the same captured camera image. Considering that, we augment the CNN inputs with the latest three estimated drone states, which include the position change and the Euler angle in the $x$-$y$-$z$ world frame, along with the linear velocity and the angular velocity in the $x$-$y$-$z$ body frame, as part of the inputs. The latest three drone states can provide information about the global trajectory $\mathcal{T}_g$ that the expert policy follows to the neural network. Our customized CNN architecture is shown in Fig. \ref{fig-neural_network}. Unlike the network proposed in \cite{sim2real}, we first explicitly extract gate-related predictions from the forward-facing camera image, which include the 2D normalized coordinates  $\vec{s}$ of the next gate's center point, the relative Euler angles $\vec{\Phi}$ of the next gate with respect to the drone, and the distance  $d$ from the drone to the next gate normalized by 10.0 m, by adding additional loss terms between these intermediate outputs and respective ground truth. These gate-related predictions are then concatenated with the latest three drone states to eventually generate the desired navigation direction $\vec{x}$ and speed $v$.

\textit{Loss function}: We design the total loss of the neural network predictions as a weighted mean square error (MSE) loss:
\begin{equation}
\begin{split}
L & = {\left\|\vec{x} - \vec{x}_{g}\right\|}^2 + \gamma_{1}(v - v_g)^2 + \gamma_{2}({\left\|\vec{s} - \vec{s}_{g}\right\|}^2) \\ & \quad + \gamma_{3}({\left\|\vec{\Phi} - \vec{\Phi}_{g}\right\|}^2) +
\gamma_{4}(d - d_g)^2,
\end{split}
\label{eq-loss}
\end{equation}
\noindent where $\vec{x}$, $v$, $\vec{s}$, $\vec{\Phi}$, and $d$ are the network outputs that have been mentioned above, whereas $\vec{x}_g$, $v_g$, $\vec{s}_g$, $\vec{\Phi}_g$, and $d_g$ denote the ground truth of them. 

Finally, the 3D goal position in the drone body frame is derived by back-projecting the 2D goal coordinates $\vec{x}$ in the image plane to get the camera projection ray and find the goal point at a distance equal to $\ell_\textrm{test}$, which is the planning length at testing time and is defined in Eq. \ref{eq-l_test}:
\begin{equation}
\ell_\textrm{test} = \textrm{min}(\ell_\textrm{max}, \textrm{max}(\ell_\textrm{min}, m_\ell \cdot v)),
\label{eq-l_test}
\end{equation}
\noindent where $\ell_\textrm{min}$ and $\ell_\textrm{max}$ are the minimum and maximum planning length, $m_\ell$ is the coefficient to let the planning length be proportional to the predicted normalized navigation speed $v$. In our case, we choose $\ell_\textrm{min} =$ 0 m,  $\ell_\textrm{max} =$ 3.0 m, $m_\ell =$ 5.0.

\subsection{Data Collection Strategy} \label{section-data_collection}

\textit{Domain transfer}: In this paper, we aim at navigating in environments with unseen photo-realistic textures while only trained with synthetic textures. Thus there is a domain shift between the training and testing data. Note that, as shown in many research \cite{rl1, sim2real_prev, transfer4}, the most important factor of successful domain transfer is the randomization of the training data, instead of its proximity to the testing data. Thus we propose a data collection policy that involves numerous randomization factors.
\begin{figure}[t]
	\setlength{\belowcaptionskip}{-10pt}
	\centering
	\subfigure[]{
		\includegraphics[width=4cm]{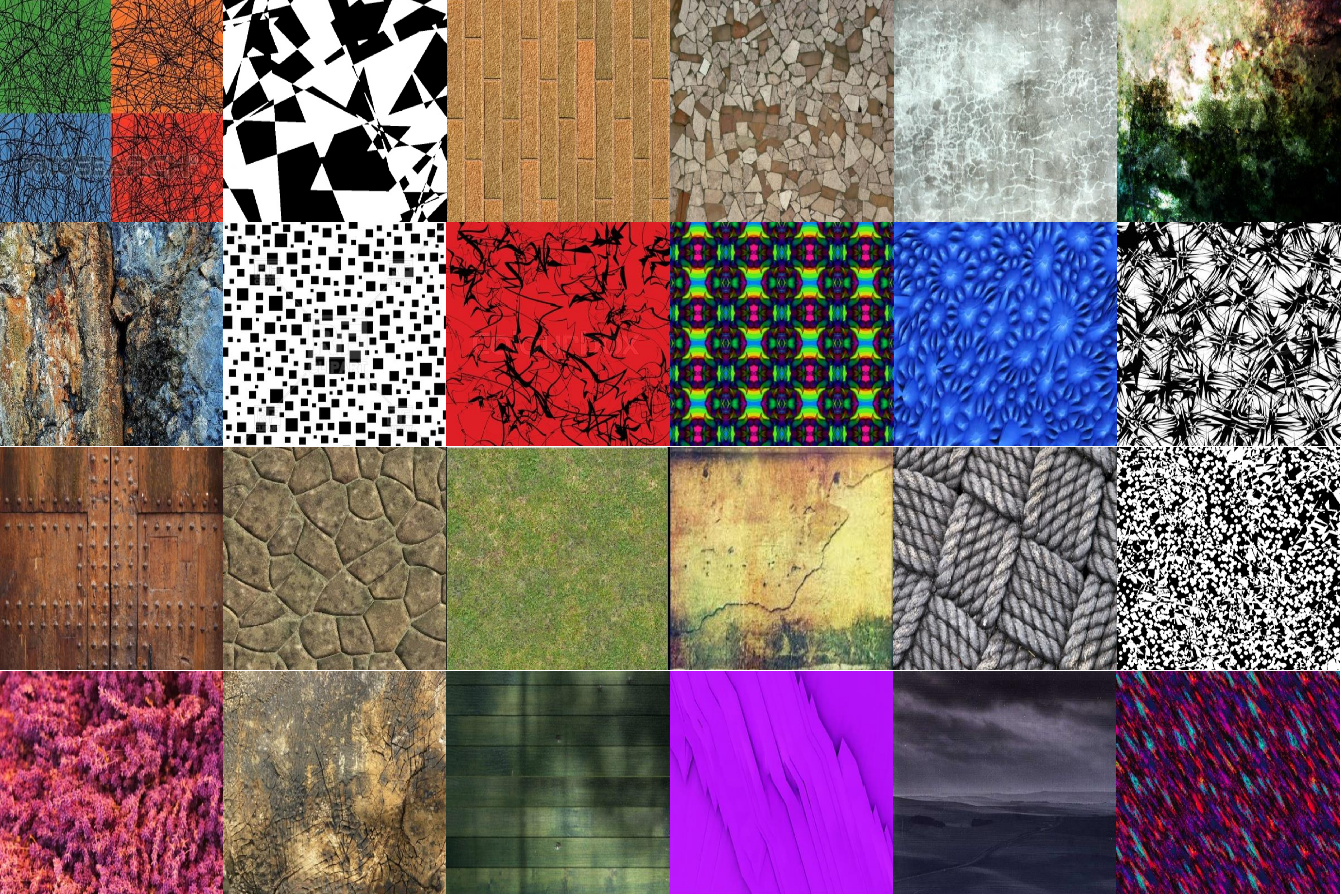}
		\label{fig-random-a}
	}
	\subfigure[]{
		\includegraphics[width=4cm]{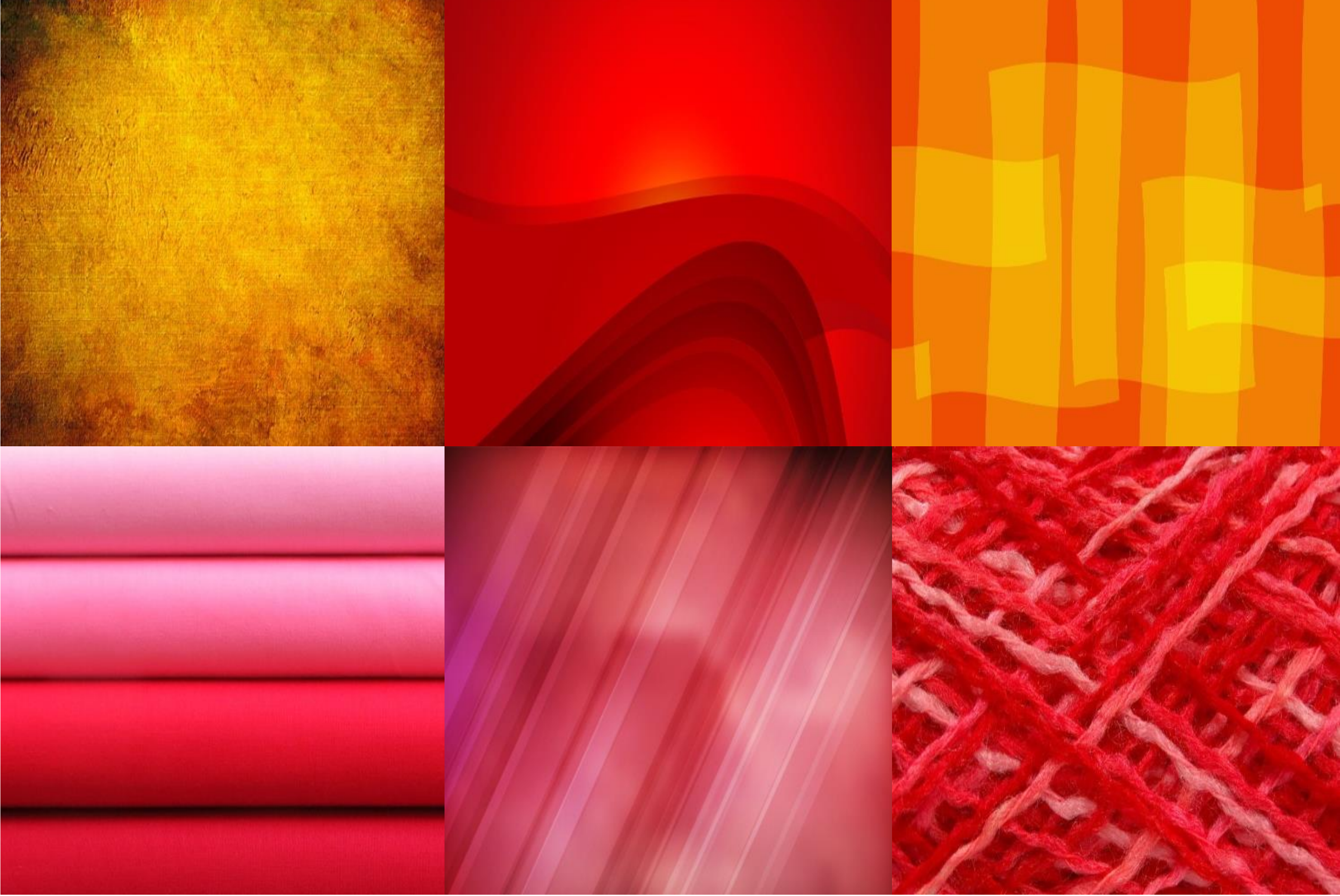}
		\label{fig-random-b}
	}
	\subfigure[]{
		\includegraphics[width=4cm]{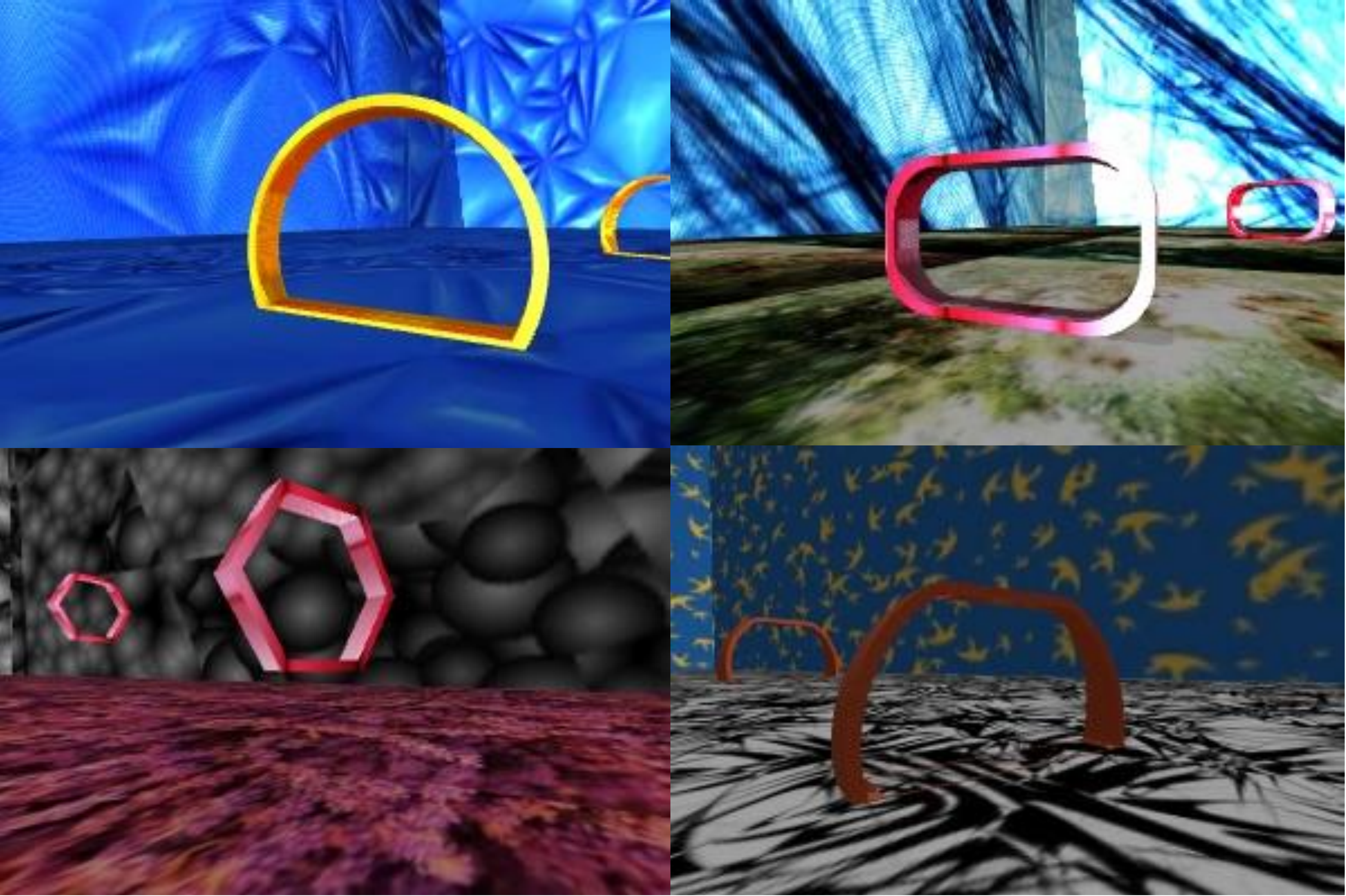}
		\label{fig-random-c}
	}
	\subfigure[]{
		\includegraphics[width=4cm]{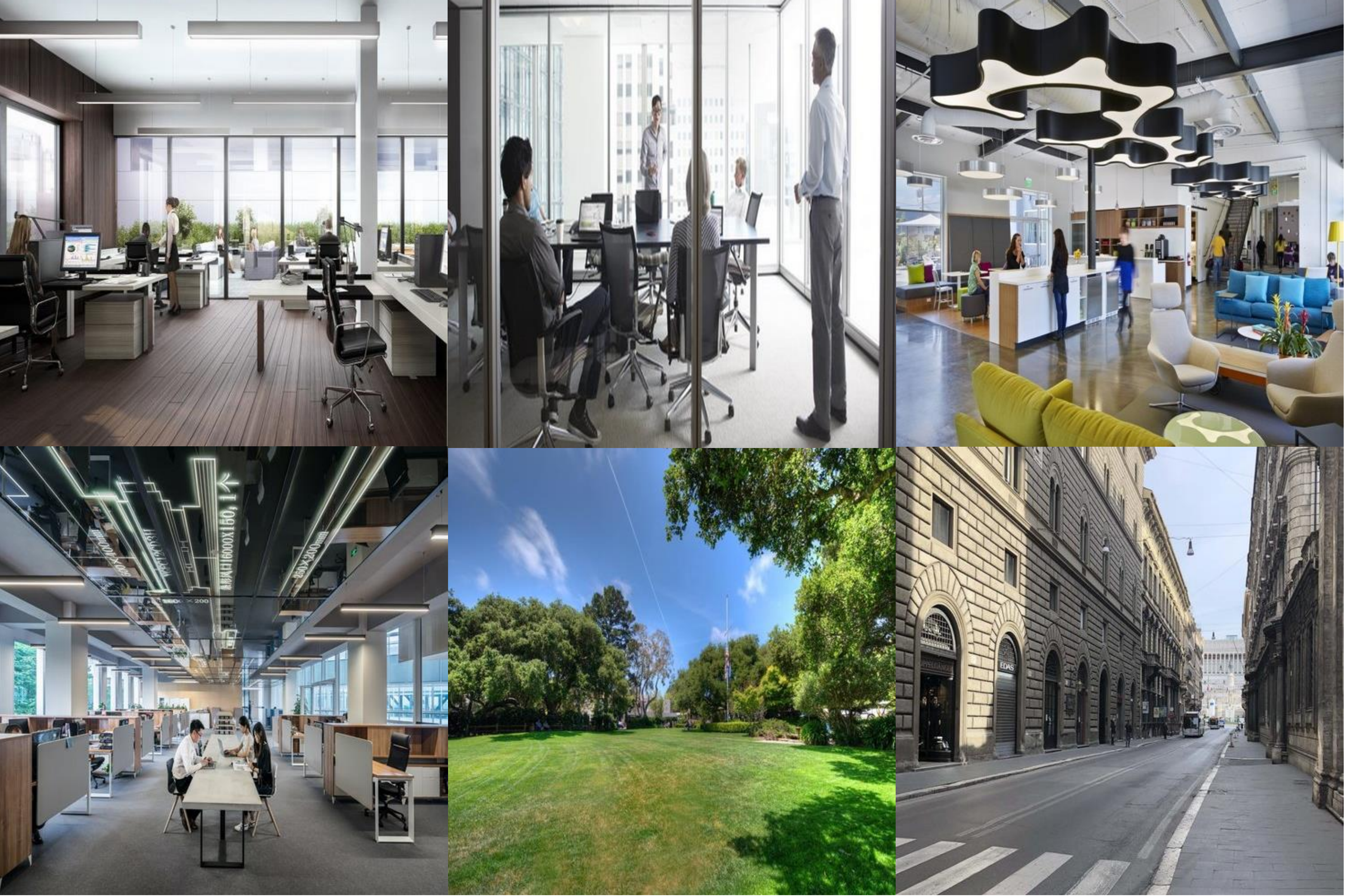}
		\label{fig-random-d}
	}
	\caption[Visual scene randomization for domain transfer.]{In order to realize domain transfer, we implement the visual scene randomization (textures of the wall and floor, shape and texture of the gates, and the illumination condition). (a) Textures of the wall and floor used during training. (b) Gate textures. (c) Several examples of traninng data illustrating our visual scene randomization. (d) Textures of the wall and floor used during testing, which were not seen during the training procedure.}
	\label{fig-random}
\end{figure}

Firstly, following the idea proposed in \cite{sim2real}, we do the visual scence randomization which randomizes:
(1) the textures of the wall and floor,
(2) the shape and texture of the gates, 
and (3) the illumination condition of the whole scene (See Fig. \ref{fig-random}).
Besides that, we also randomize the following factors for every new data collection epoch:
i) the gate positions, 
ii) the navigation direction, and 
iii) the maximum velocity for the global and local trajectories.    
For i), the track layout we choose for experiments is primarily a circle composed of seven gates, but the position for each gate is randomized for each experiment. For ii), a random Boolean number is generated to determine whether the expert policy flies clockwise or counterclockwise. As for iii), we generate the maximum velocity $v_\textrm{max}$ uniformly from 4.0 to 8.0 m/s.

\textit{Dataset Aggregation (DAgger) policy}: In this paper, we train the perception module to imitate the expert policy. However, a key drawback of imitation learning is the distribution mismatch between the training and testing data. In our case, the expert policy would maneuver the drone close to the reference global trajectory $\mathcal{T}_g$, thus missing the data for scenarios in which the drone is far away from $\mathcal{T}_g$. The Dataset Aggregation (DAgger) policy \cite{DAgger} is thus usually used in imitation learning to alleviate this impact, of which the key concept is to execute the partially trained network to maneuver the robots but asking the expert policy to supervise those scenarios. As shown in Fig. \ref{fig-intro_DAgger}, we use a variant of DAgger policy during data collection, which uses the partially trained network to maneuver the drone but switches to the expert policy to recover the drone when it flies far away from the global trajectory (defined as a margin $\epsilon$). Whether is the neural network or the expert policy being executed, we always ask the expert policy to generate its decisions in those scenarios and store them in the training data. The details are discussed below.

We first collect the testing data by executing the expert policy in environments with photo-realistic textures as shown in Fig. \ref{fig-random-d} for 100 trials, 40 seconds each. 

As for the training data, only synthetic textures as shown in Fig. \ref{fig-random-a} are used. The training and data collection procedure is as follows: 
\begin{enumerate}[1)\indent]
	\item Collect training data using only the expert policy for 150 trials, 40 seconds each. After that, train the neural network for 100 epochs. 
	\item Set the margin $\epsilon = 0.5$ m, use the modified DAgger policy mentioned above with the partially trained neural network and collect augmented data for 150 trials. Train the neural network from scratch on all the accumulated data for 100 epochs.
	\item Set $\epsilon = 1.0$ m, and repeat step 2.
	\item Set $\epsilon = 1.5$ m, and repeat step 2. After that, stop since further repeat does not improve the performance and the performance of our final trained network is satisfactory.
\end{enumerate}

Setting the margin $\epsilon$ to a large value would allow collecting more diverse data. Nevertheless, with a poorly-performed partially-trained neural network, the expert policy may not be able to recover the drone if the margin $\epsilon$ is set to be too large. As a result, we gradually increase the margin $\epsilon$ as the trained neural network performs better and better. Besides, surprisingly, we have found that after each step of data augmentation, training the network from scratch outperforms the one starting from the pre-trained weights, which might be explained by it being stuck at a local minimum and detailed comparison will be shown in Section \ref{section-training_from_scratch}. 

\textit{Dynamic environments}: The aforementioned training data collecting procedure is limited to static track layouts since our expert policy requires a pre-computed global trajectory that passes through all the gates. Besides, calculating a new global trajectory at every step would be too time-consuming for the expert policy to perform well. So how can we train a perception module that would be able to handle dynamic environments? Following the idea proposed in \cite{sim2real}, during experiments, we have found that training on multiple static environments is sufficient for the trained network to navigate in dynamic environments at testing time. This simple approach which randomizes the track layout during training time enables the trained network's generalization to dynamic tracks and thus improves its robustness.

\section{EXPERIMENTS}
A simulator called RotorS \cite{rotors}, which models the dynamics of the drones with high fidelity and is developed with Gazebo, is used throughout the experiments for both training and testing. The flying space is bounded by the walls and floor whose textures are randomized through the experiments as mentioned in Section \ref{section-data_collection}.

The easiest track layout in our experiments is a circular track composed of seven gates placed at the same height (0.0 m), whereas the difficult track perturbs the gate positions with a maximum distance of 2 m before the drone starts. The dynamic track, on the other hand, continuously moves the gate while the drone flies, which adds more difficulty for the racing drones.

\begin{figure*}[t]
	\centering
	\subfigure[]{
		\includegraphics[width=8.5cm]{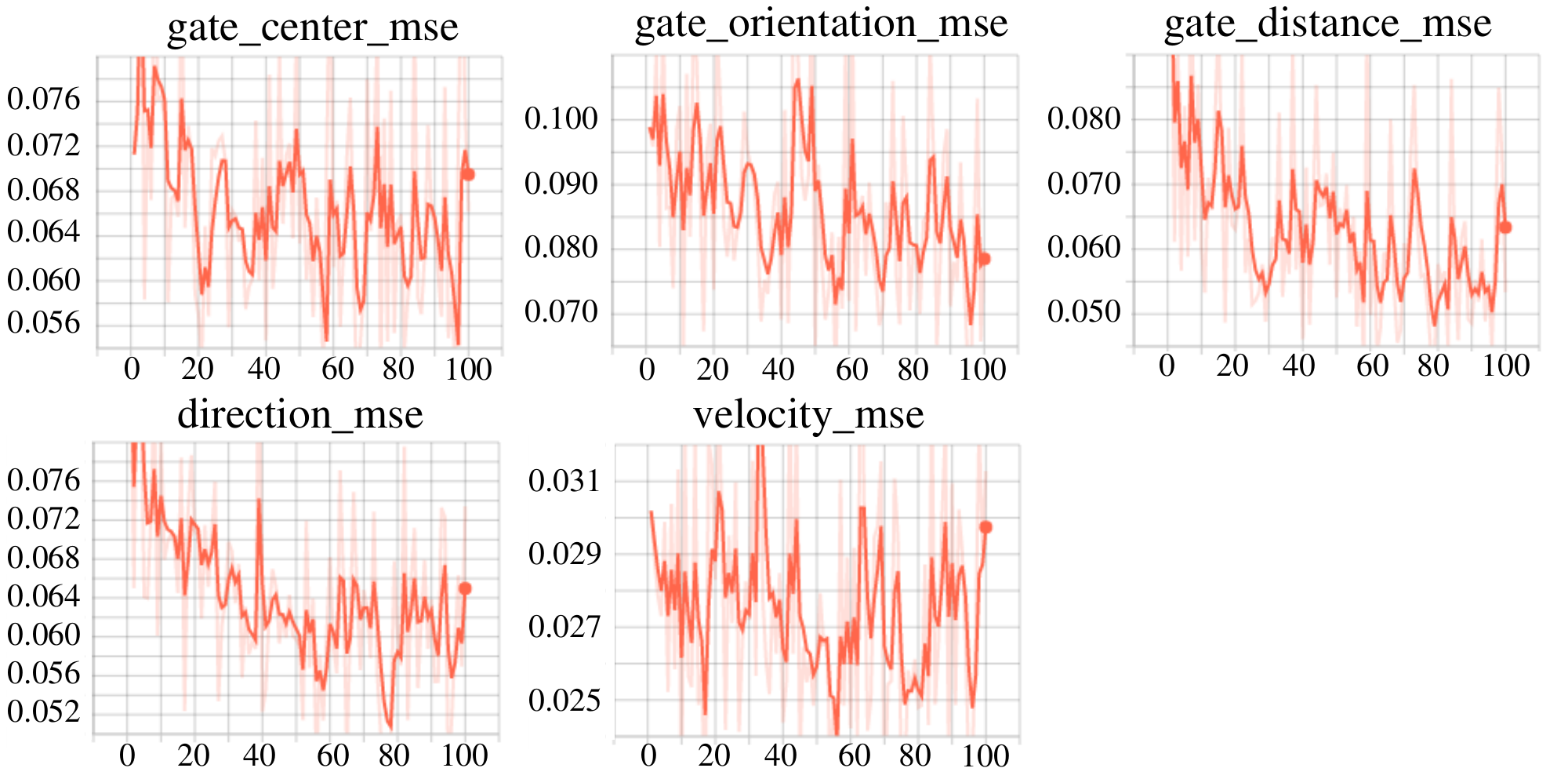}
		\label{fig-training_procedure-with}
	}
	\subfigure[]{
		\includegraphics[width=8.5cm]{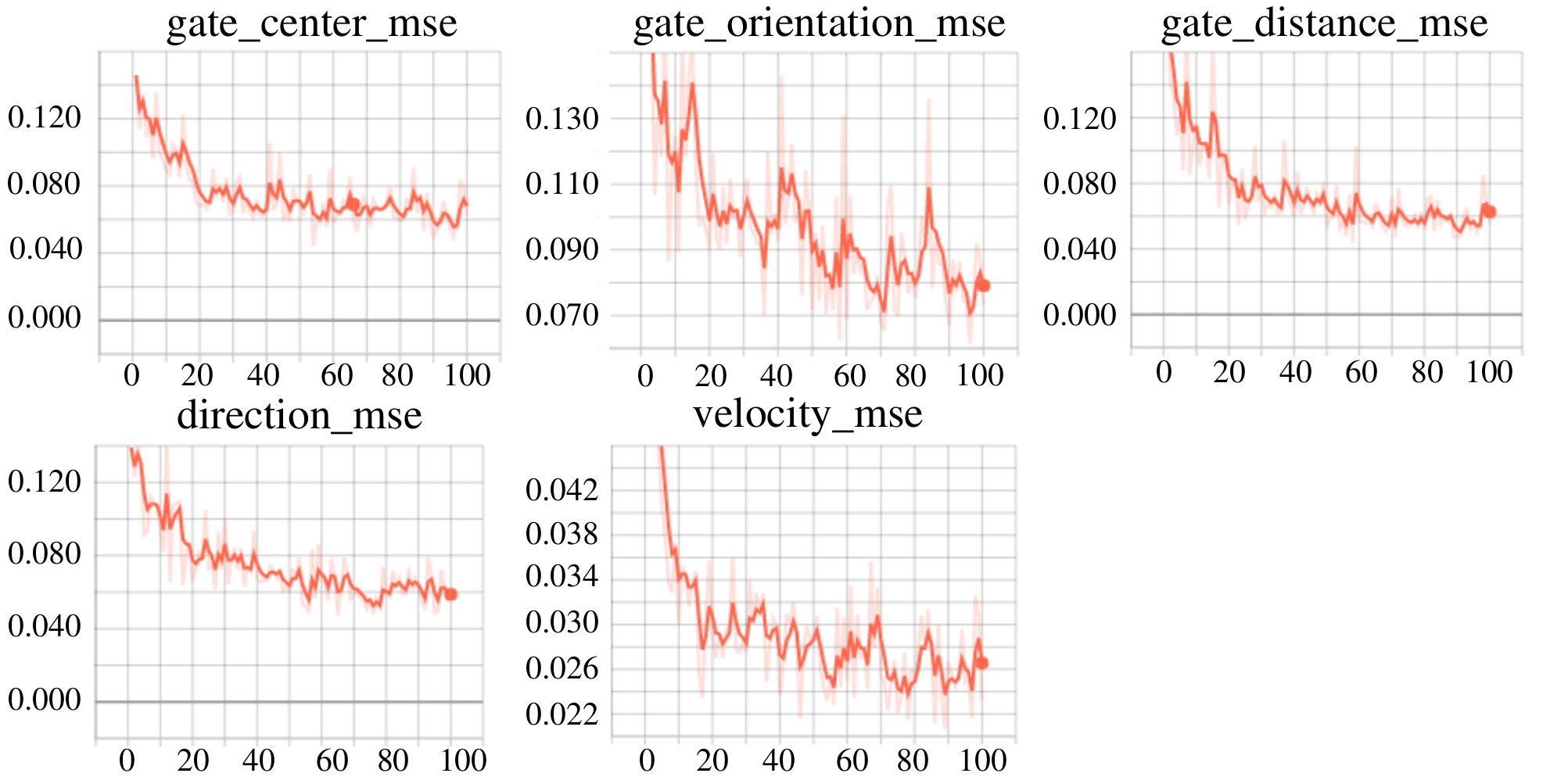}
		\label{fig-training_procedure-without}
	}
	\caption[Comparison of training procedure choices (with warming-up VS without warming-up).]{Test loss of the training process after data augmentation (a) when using pre-trained weights, and (b) when training from scratch. The x-axis in all plots represents training epochs.} 
	\label{fig-training_procedure}
\end{figure*}
In the following sections, we first find the optimal network architecture and the optimal loss function coefficients during training. After that, we validate our choice of training from scratch as described in our modified DAgger policy (see Section \ref{section-data_collection}). We then compare our proposed method's overall navigation performance to the state-of-the-art method in \cite{sim2real}. Finally, additional experiments have been done to figure out which randomized factors contribute more to the success of the domain transfer.  

\subsection{Neural Network Capacity}
Given the limited computation resource on racing drones' onboard processing units, it is crucial to find an optimal neural network capacity (depth) that achieves satisfactory performance with short inference time. Similar to \cite{sim2real}, we have done comparisons between networks with different capacity, and have chosen the network with the optimal capacity as our customized neural network structure (see Fig. \ref{fig-neural_network}).

\begin{table}[t]
	\setlength{\abovedisplayskip}{0.cm} 
	\setlength{\belowdisplayskip}{-0.cm}
	\vspace{-0.2cm}
	\caption[Comparison of the trained neural networks in terms of their testing loss and closed-loop performance under different coefficient settings in loss function.]{Comparison between neural networks trained under different coefficient choices in loss function.}
	\label{tab-coefficient}
	\begin{center}
		\begin{tabular} {cccc|cc}
			\hline\hline
			$\gamma_1$ & $\gamma_2$ & $\gamma_3$ & $\gamma_4$ & Test Loss & Completion Rate [\%]\\
			\hline
			0.1 & 1 & 0.2 & 0.2 & 0.272 & 100\\
			1 & 1 & 0.2 & 0.2 & 0.274 & 100\\
			10 & 1 & 0.2 & 0.2 & 0.291 & 100\\
			100 & 1 & 0.2 & 0.2 & 0.351 & 82\\
			0.1 & 1 & 0.1 & 0.1 & 0.280 & 100\\
			0.1 & 0.05 & 0.1 & 0.1 & 0.302 & 85\\
			0.1 & 0.5 & 0.1 & 0.1 & 0.295 & 100\\
			\hline\hline
		\end{tabular}
	\end{center}
\end{table}

\subsection{Coefficient Choice for the Loss Function}
As shown in Table \ref{tab-coefficient}, various choices of the loss function coefficients defined in Eq. \ref{eq-loss} has been compared to find an appropriate setting. Note that, when calculating the test loss, we use the same coefficients for fair comparison ($\gamma_1 =$ 0.1, $\gamma_2 =$ 1.0, $\gamma_3 =$ 0.2, $\gamma_4 =$ 0.2). The completion rate on the track is defined as proportional to the number of passed gates, which 100 $\%$ indicates finishing five loops without a crash. 

According to the results shown in Table \ref{tab-coefficient}, the trained policy's performance is not very sensitive to the selection of the coefficients. Still, there are several interesting findings. First of all, giving a too large coefficient to one of the loss terms (see the fourth row of Table \ref{tab-coefficient} where $\gamma_1 =$ 100) might undermine the trained policy's performance. Secondly, the coefficient for the loss term of the 2D image coordinates of the next gate's center position ($\gamma_2$), which are part of our neural network's intermediate outputs, should not be set too small (see the second from last row of Table \ref{tab-coefficient}, where $\gamma_2 =$ 0.05). This indicates that our gate-related intermediate outputs make the trained policy perform better, which shows the advantage of our network architecture compared to the one proposed in \cite{sim2real}. Based on Table \ref{tab-coefficient}, we choose $\gamma_1 =$ 0.1, $\gamma_2 =$ 1, $\gamma_3 =$ 0.2, $\gamma_4 =$ 0.2 as the optimal coefficients.

\subsection{Procedure of Data Collection and Training} \label{section-training_from_scratch}
A core step of our modified DAgger (Dataset Aggregation) strategy is to train the neural network from scratch after collecting augmented data, which is in contrast to the procedure in \cite{sim2real} where the neural network uses the previously trained weights as the initial weights and is then fine-tuned on all the accumulated data.

As shown in Fig. \ref{fig-training_procedure}, training from scratch shows much more stable and better results, whereas training procedure with the pre-trained weights heavily fluctuates during the training process. This may be explained as the neural network being stuck in a local minimum since the pre-trained weights were obtained from the data before data augmentation. 

\begin{figure*}[t]
	\centering
	\subfigure[]{
		\includegraphics[width=5cm]{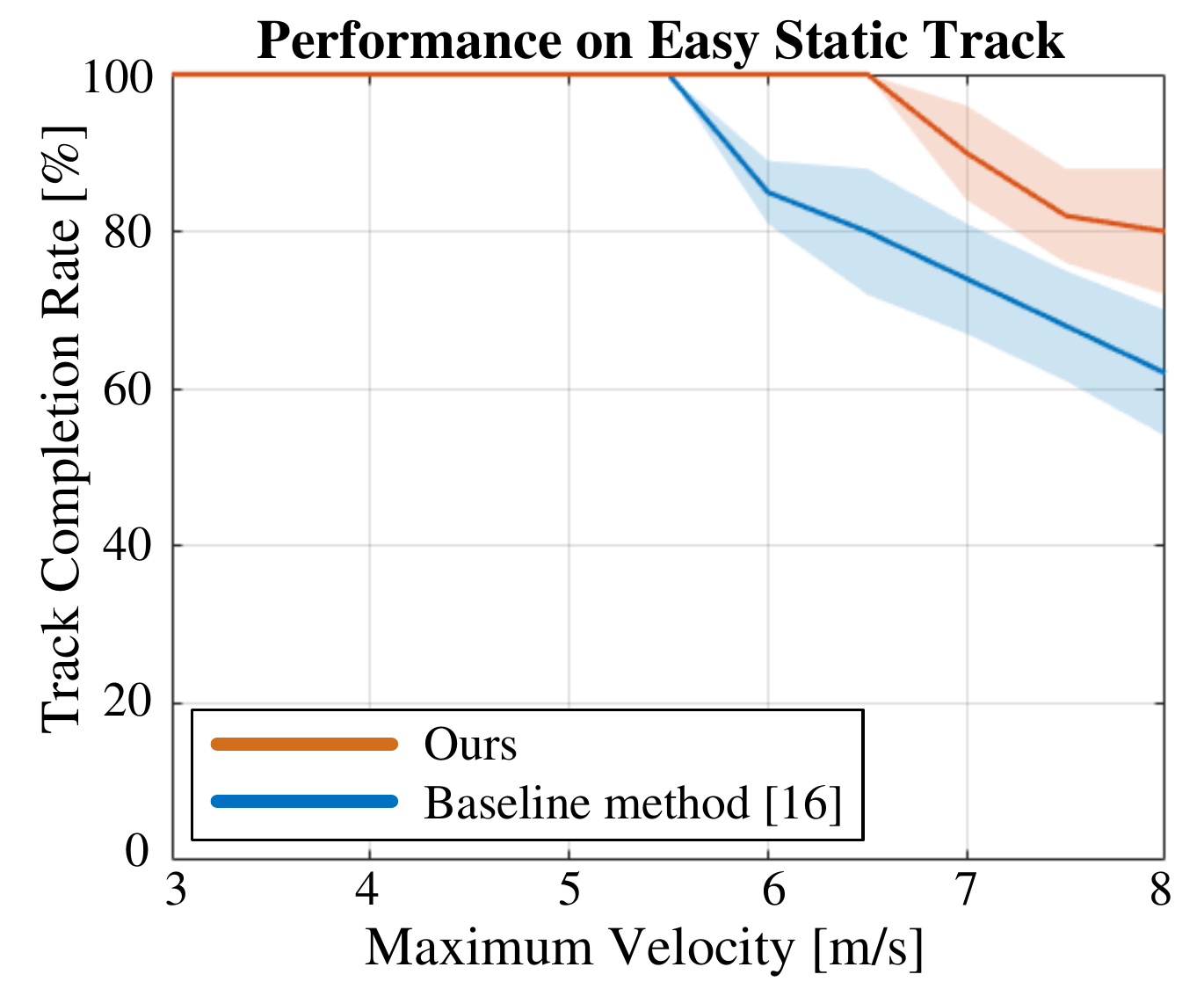}
		\label{fig-performance_evaluation-a}
	}
	\subfigure[]{
		\includegraphics[width=5cm]{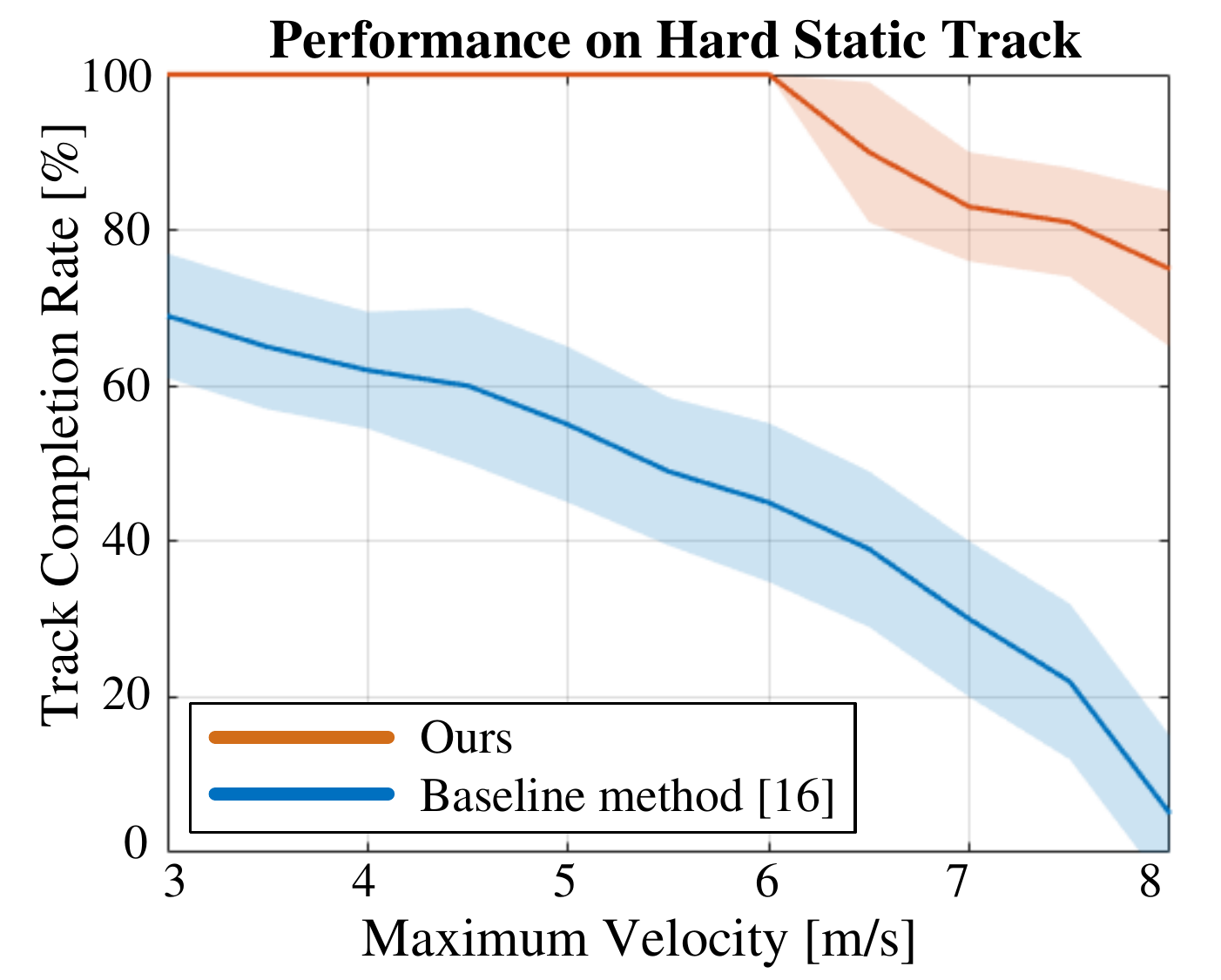}
		\label{fig-performance_evaluation-b}
	}
	\subfigure[]{
		\includegraphics[width=5cm]{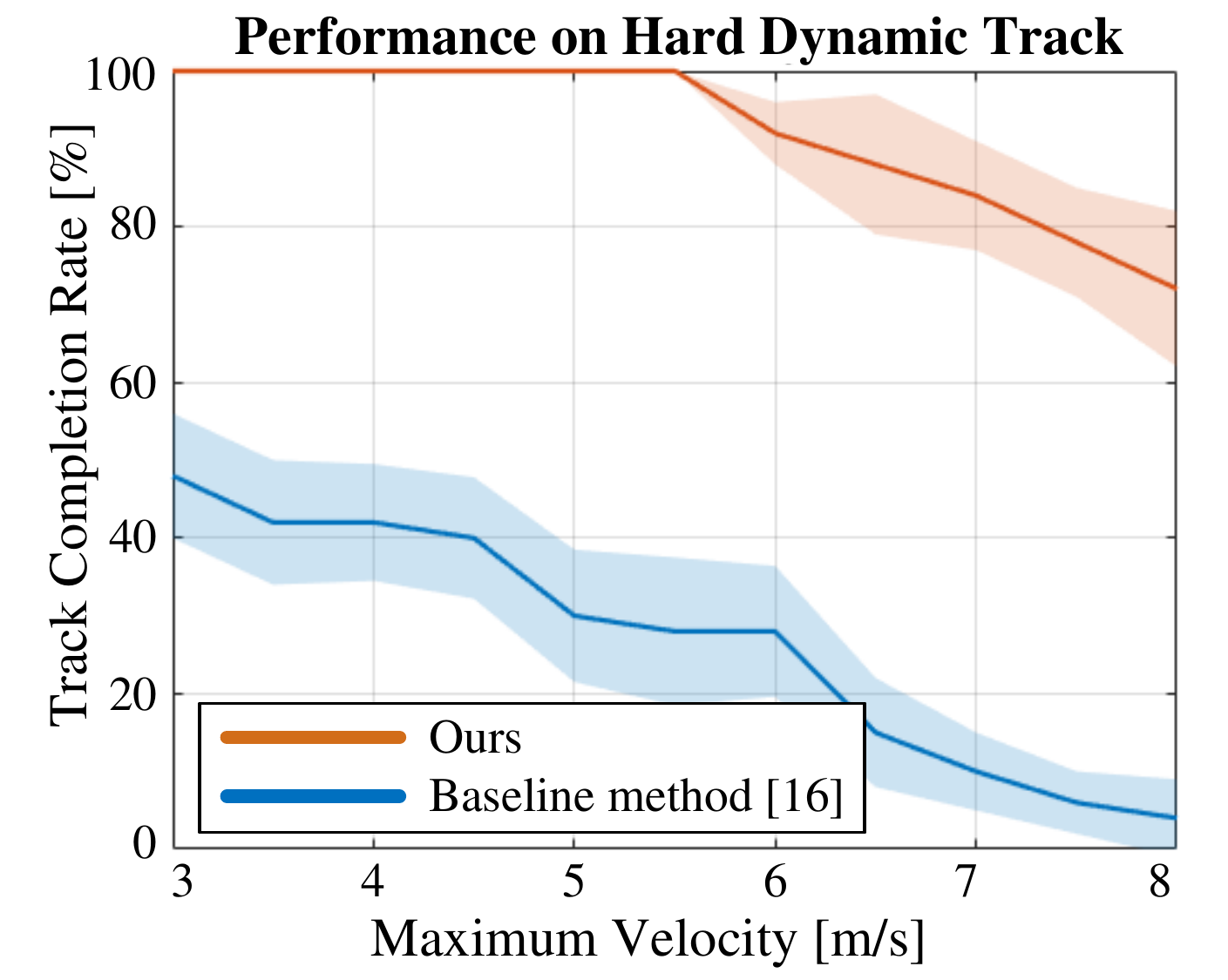}
		\label{fig-performance_evaluation-c}
	}
	\caption[Performance of the proposed method compared to the state-of-the-art method.]{Performance comparison between our proposed method and a baseline state-of-the-art method \cite{sim2real}. The lines represent mean values of the completion rate, while the shaded areas indicate their standard deviation.}
	\label{fig-performance_evaluation}
\end{figure*}

\begin{figure}[t]
	\centering
	\subfigure[]{
		\includegraphics[width=4cm]{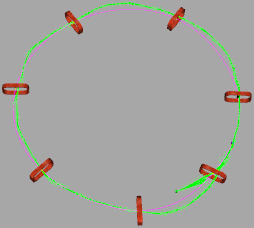}
		\label{fig-direct_no}
	}
	\subfigure[]{
		\includegraphics[width=4cm]{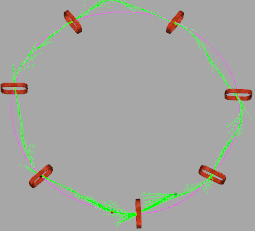}
		\label{fig-direct_yes}
	}
	\caption[Different navigation behavior using a single trained neural network]{Different navigation behavior using a single trained neural network (both in counterclockwise). (a) use the original final output of the neural network as the navigation direction (b) use the intermediate output of the neural network which aims at the center position of the next gate as the navigation direction}
	\label{fig-direct}
\end{figure}
\subsection{Performance Evaluation Comparing to Previous Methods}
The overall performance of our trained policy is compared with the baseline state-of-the-art method proposed in \cite{sim2real} on various tracks with different complexity. For each different experimental setup, five independent testing trials have been done to calculate the mean value and the standard deviation of the track completion rate.

As shown in Fig. \ref{fig-performance_evaluation}, our proposed method outperforms the baseline state-of-the-art method \cite{sim2real} especially on more difficult track layouts, where the gates could be irregularly placed inside the flying space or even keep moving at testing time. 

Another advantage of our proposed method is that we can easily switch our trained policy's behavior by alternatively using the CNN prediction of the next gate's center point as our navigation direction (see Fig. \ref{fig-neural_network}). In this way, the drone always flies straight to the gate (see Fig. \ref{fig-direct_yes}). This kind of behavior is preferable when the drone flies at high speed or the distance between gates is very close such that there is not much time for the navigation policy to respond. However, when there is a significant orientation difference between the gate and the drone, flying straightly to the center point of the gate would end up crashing. In conclusion, our customized neural network gives us two options for the navigation policy with a single trained network.

\subsection{Analysis of Factors that Contribute to Domain Transfer}
Given various factors that are randomized when collecting data as mentioned in Section \ref{section-data_collection}, it is useful and insightful to figure out which factor contributes more to the success of the domain transfer. To do so, we disable the randomization for each factor respectively during collecting data and compare their respective loss on our photorealistic testing data.

The testing data we use is collected in the difficult static track layout, in which the gate positions are randomly perturbed and so is the flying direction. As a result, to make the trained policy be able to fly in difficult tracks in both directions,  the randomization for gate positions and navigation direciton is indispensable. In our experiment, the trained policy without these two factors cannot pass a single gate in difficult static tracks and the testing losses are also too large to put in the same figure with other factors. 

\begin{figure}[t]
	\centerline{\includegraphics[width=9cm]{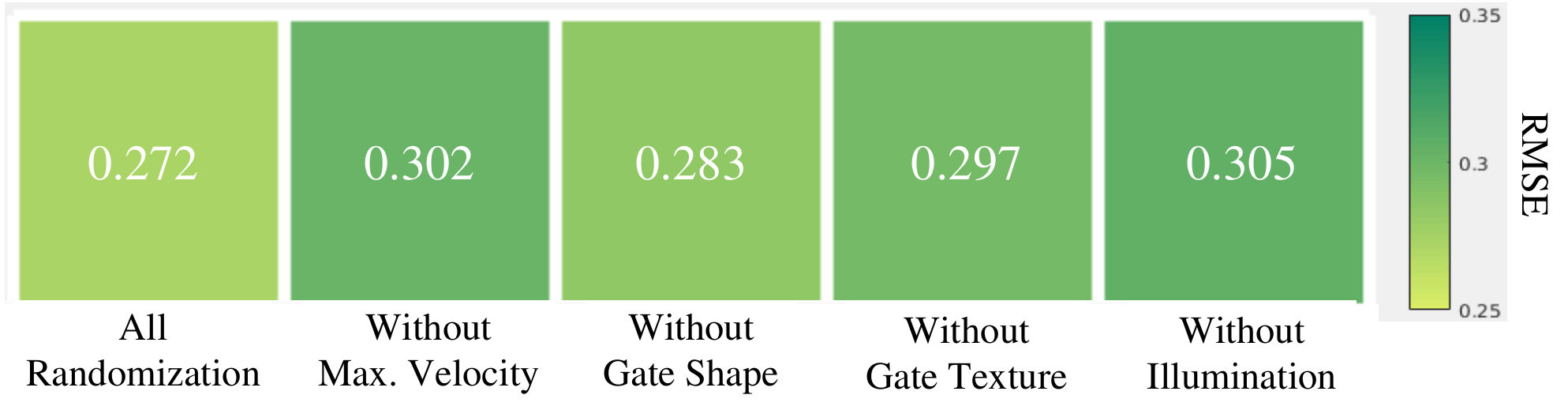}}
	\caption[Contribution of the randomization factors to domain transfer]{Contribution of the randomization factors to domain transfer. The metric here is the RMSE (root-mean-square-error) loss on the testing data, for which lower means better performance.} 
	\label{fig-random_factor}
\end{figure}

The background (wall and floor) texture randomization is well known to be critical for transfer \cite{sim2real}, thus we omit the comparison of turning this factor on and off. The factors that cause significant performance drops if not being randomized during collecting data are those that contribute heavily to the transfer. As shown in Fig. \ref{fig-random_factor}, the randomization of the maximum velocity and the illumination condition matter the most in the trained neural network's performance, while the effect of gate shape randomization is not significant. 

\section{CONCLUSIONS}
In this paper, we have proposed a novel vision-based approach for autonomous racing drones. Comparing to the method in \cite{sim2real}, we add the latest three drone states along with the current camera image as the inputs of our customized neural network, which serves as the perception module and generates the desired navigation direction and speed that are then executed by a state-of-the-art planner and a controller. Besides, the extensive amount of randomization during data collection and our modified DAgger (Dataset Aggregation) policy makes our trained network robust to condition changes. While purely trained in simulation environments with synthetic textures, our trained network successfully operates with unseen photorealistic textures during testing. Overall, our trained policy can navigate through hard dynamic tracks at a maximum velocity of up to 6.0 m/s without a crash, whereas the baseline method \cite{sim2real} could barely pass one or two gates. Besides, with a single trained network, we have another policy behavior available by instead using the predicted 2D image coordinates of the next gate's center point as the navigation direction, which makes the drone always fly straight to the next gate.

\section*{ACKNOWLEDGEMENT}
This work was supported by the National Research Foundation of Korea (NRF) grant funded by the Korea government (MSIT) (No. 2021R1A2C2010585) and by the Institute for Information $\&$ Communications Technology Planning $\&$ Evaluation (IITP) grant funded by the Korea government (MSIT) (No. 2019-0-01396).










\end{document}